\documentclass{INTERSPEECH2023}

\usepackage{color, soul, multirow}
\usepackage{subfig}
\usepackage{bm}
\usepackage{verbatim}
\usepackage{amssymb}
\usepackage{pifont}
\usepackage{float}
\usepackage{booktabs}
\usepackage{pifont}
\usepackage{tikz}
\usepackage{lipsum}  
\interspeechcameraready

\title{Modality Confidence Aware Training \\
for Robust End-to-End Spoken Language Understanding}
\name{Suyoun Kim$^1$, Akshat Shrivastava$^1$, Duc Le$^2$, Ju Lin$^1$, Ozlem Kalinli$^1$, Michael L. Seltzer$^1$\thanks{The work is performed during Duc Le is at Meta.}}
\address{
  $^1$Meta, USA\\
  $^2$TikTok, USA}
\email{suyounkim@meta.com}

\begin{document}

\maketitle
 
\begin{abstract}
    End-to-end (E2E) spoken language understanding (SLU) systems that generate a semantic parse from speech have become more promising recently. This approach uses a single model that utilizes audio and text representations from pre-trained speech recognition models (ASR), and outperforms traditional pipeline SLU systems in on-device streaming scenarios. However, E2E SLU systems still show weakness when text representation quality is low due to ASR transcription errors. To overcome this issue, we propose a novel E2E SLU system that enhances robustness to ASR errors by fusing audio and text representations based on the estimated modality confidence of ASR hypotheses. We introduce two novel techniques: 1) an effective method to encode the quality of ASR hypotheses and 2) an effective approach to integrate them into E2E SLU models. We show accuracy improvements on STOP dataset and share the analysis to demonstrate the effectiveness of our approach.
\end{abstract}

\section{Introduction}
\label{sec:intro}
    As voice-driven device interfaces continue to gain popularity in conversational AI, spoken language understanding (SLU) systems that generate semantic parse from speech are becoming increasingly important. Traditional SLU systems are typically composed of two separate components, including automatic speech recognition (ASR) and natural language understanding (NLU). The ASR component generates transcriptions from speech, while the NLU component generates semantic parse from the ASR's output hypotheses. While the pipeline approach allows for individual development of ASR and NLU components, it also has limitations such as being vulnerable to error propagation from ASR to NLU, having limited acoustic information which reduces NLU accuracy, and a lack of parameter sharing that hinders on-device deployment of SLU. 

    An alternative End-to-End (E2E) approach \cite{Haghani18SLU, serdyuk2018endtoend, Potdar2021ASE, Radfar2021FANSFA, Wang2021Speech2SlotAE, 2020NeuralInterface, raju2021end, arora2022two, xu2022introducing} that directly converts speech into semantics has been recently introduced to address these limitations of pipeline approach. The E2E approach has shown improved performance in domain or intent prediction and slot tagging tasks. However, there are only a few studies on E2E SLU in resource-constrained environments \cite{arora2022two, le2022deliberation, desot2022end}. 
    
    More recently, the paper \cite{arora2022two, le2022deliberation} proposed a deliberation-based approach to E2E SLU inspired by two-pass E2E ASR \cite{sainath2019two, li2020parallel, xu2022rescorebert, kim2022joint}. Particularly, \cite{le2022deliberation} shows promising results in resource-constrained on-device environments. However, we found that the model can still experience limitations in cases where the quality of text representation is compromised due to transcription errors from the ASR component. This is especially problematic in on-device streaming use cases where ASR performance is often poor due to limited resources and contextual information.

    \begin{figure}[t]
    \begin{minipage}[b]{1.0\linewidth}
      \centering
      \centerline{\includegraphics[width=4.5cm]{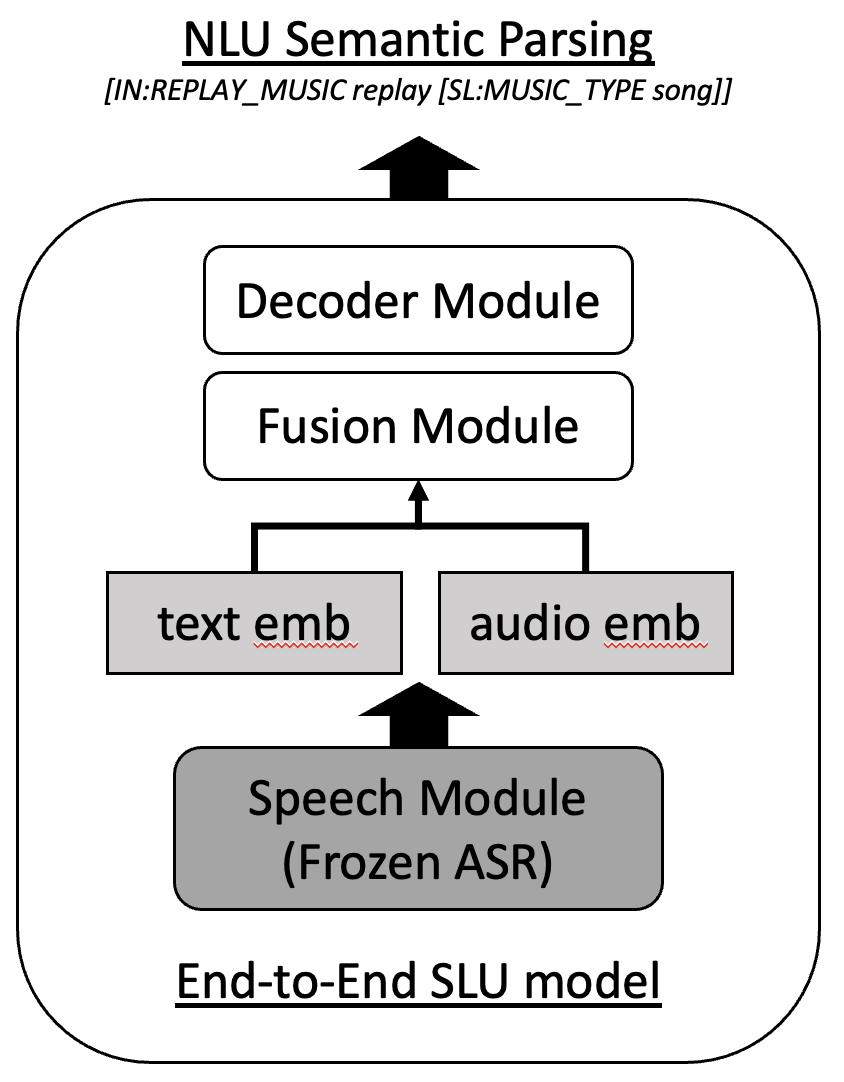}}
    \end{minipage}
    \caption{The overall architecture of End-to-End Spoken Language Understanding model.}
    \label{fig:base_slu}
    \end{figure}

    In this work, we propose a new E2E SLU system that improves its robustness to ASR errors by fusing audio and text representations wisely, taking into account the estimated confidence scores of each modality representation. Our approach distinguishes itself from previous studies \cite{le2022deliberation} by its ability to dynamically shift focus towards the audio representation in cases where the ASR output hypotheses contain errors. We present two novel techniques to improve E2E SLU models: 1) a method to encode ASR hypothesis quality and 2) an effective method to integrate these quality information into E2E SLU models. We show accuracy improvements on STOP dataset \cite{stop2022} in the on-device streaming scenario and share the analysis to demonstrate the effectiveness of our approach.

\section{Modality Confidence Aware Training}
\label{sec:mcat} 

    \subsection{Deliberation NLU component in E2E SLU}
    \label{sec:deliberation}
    
    In our proposed Modality Confidence Aware Training (MCAT), we extend the deliberation model \cite{le2022deliberation}, which is one of the latest on-device streaming E2E SLU techniques. Unlike pipeline SLU systems with separate ASR and NLU models, the deliberation-based E2E SLU models optimize ASR and NLU components jointly. For on-device streaming, the Recurrent Neural Network Transducer (RNNT) \cite{graves2012sequence, Prabhavalkar17, battenberg2017exploring, he2019streaming, li2019improving, kim2021improved} is used as the ASR component. The ASR component is trained separately and then frozen to maintain transcription accuracy. This is because fine-tuning the ASR model on SLU data (which is relatively smaller than ASR training data) may negatively impact its performance on out-of-domain test cases, such as long-form transcription tasks.

    The vanilla deliberation-based NLU component consists of two main modules: (1) \texttt{Fusion}, and (2) \texttt{Decoder} module. Rather than using the RNNT's final hypotheses directly, the \texttt{Fusion} module in NLU component takes the intermediate audio and text representation, such as text embeddings $e_{\text{txt}}^{1:U}$ from the \texttt{Predictor} of the frozen RNNT and audio embeddings $e_{\text{aud}}^{1:T}$ from the \texttt{Encoder} of the frozen RNNT. By using Multi-Head Attention($\texttt{MHA}_\texttt{fusion}$), the \texttt{Fusion} module generates the fused feature $e_{\text{fused}}^{1:U}$ as follows:
    \begin{align}
        e_{\text{fused}} &= \texttt{MHA}_\texttt{fusion}(e_{\text{txt}}, e_{\text{aud}}, e_{\text{aud}})
    \end{align}
    Note that we used $e_{\text{txt}}$ as a query, $e_{\text{aud}}$ as a key and value in \texttt{MHA}. 

    Given the fused feature $e_{\text{fused}}$ from the \texttt{Fusion} module, the \texttt{Decoder} module generates the final target semantic token distributions ($o^{1:V}$) by using transformer-based decoder with \textit{pointer-generator} technique \cite{see2017get, decoupled}. At each output time step $v$, (1) the probability of generating a new semantic token ($g_v$) and (2) the probability of copying a token ($c_v$) are computed and combined together based on a mixing probability $P_\texttt{copy}$, then the final output token distribution ($o_v$) is computed. The copying probability ($c_v$) is computed from the decoder state, and the generating probability ($g_v$) is computed from the $\texttt{MHA}_\texttt{dec}$. 

    \begin{align}
        g_v &= \texttt{Softmax}(\texttt{Linear}(d_v)) \\ 
        c_v, a_v &= \texttt{MHA}_\texttt{dec}(d_v, e_{\text{fused}}, e_{\text{fused}})  \\
        P_\text{copy} &= \sigma(\texttt{Linear}([d_v, c_v])) \\
        o_v &= \texttt{Softmax}(P_\texttt{copy} \cdot c_v + (1-P_\texttt{copy}) \cdot g_v)
    \end{align}

    Figure \ref{fig:base_slu} describes the overall architecture of the deliberation-based E2E SLU systems \cite{le2022deliberation}. While the deliberation E2E SLU models has been shown to mitigate ASR error propagation by leveraging both audio and text representations, we observed that relying solely on audio representation can yield better results when the ASR hypothesis contains errors. This observation suggests that effectively incorporating both audio and text modalities could further improve NLU performance, especially in cases where the text representation is unreliable due to ASR hypothesis errors. 
    
    \subsection{Integration of Modality Confidence Score}
    \label{sec:integration}
    
    \begin{figure}[t]
    \begin{minipage}[b]{1.0\linewidth}
      \centering
      \centerline{\includegraphics[width=8cm]{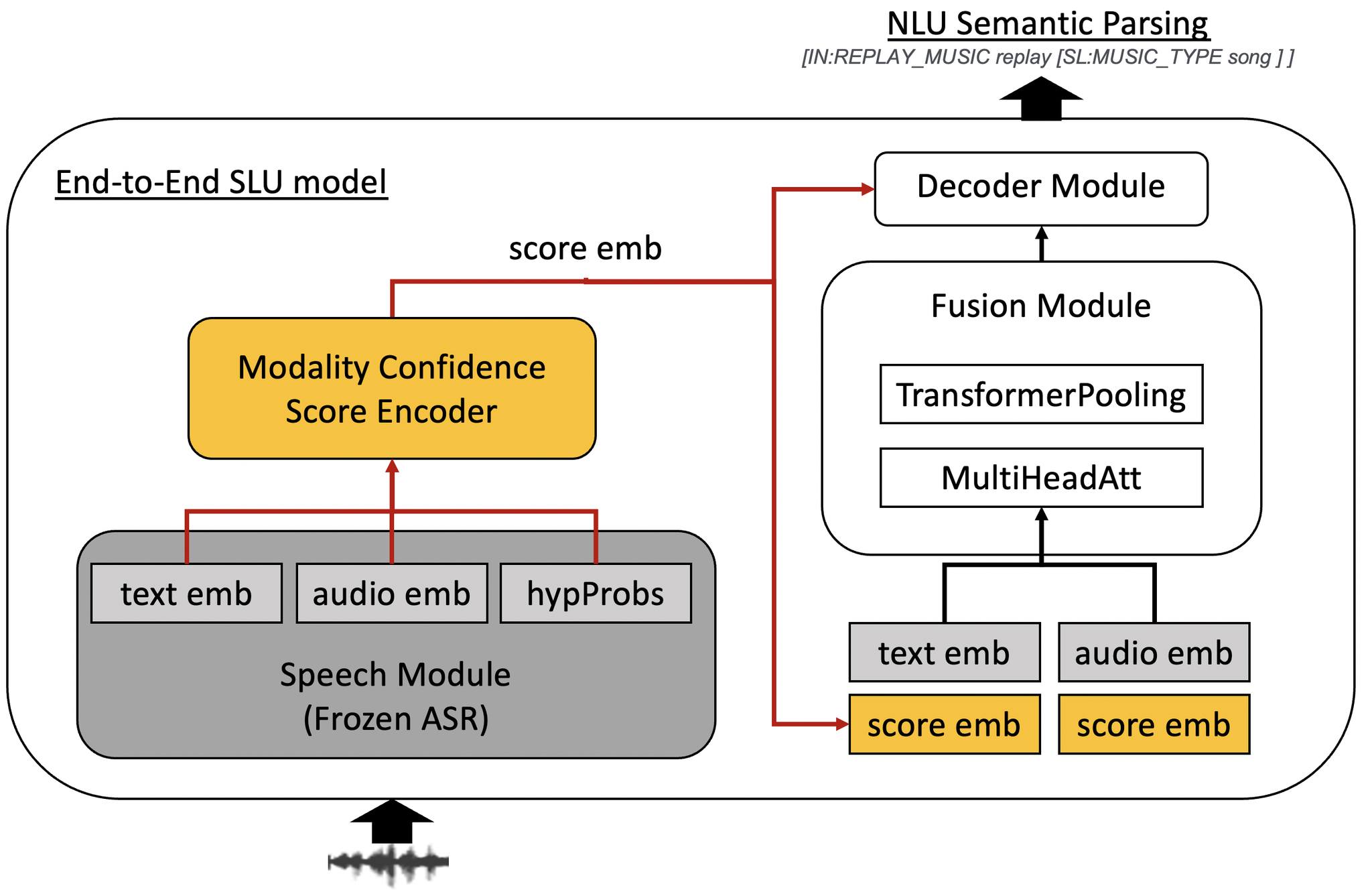}}
    \end{minipage}
    \caption{The overall architecture of our proposed Modality Confidence Aware Training (MCAT).}
    \label{fig:attention}
    \end{figure}
    
    The main idea of our proposed approach, MCAT, is to build the E2E SLU model that can wisely combine audio and text representation based on the confidence level of each modality. For example, when the output of the ASR component is accurate, the NLU component will rely more on the text representation, while the output of the ASR component contains errors, the NLU component will rely more on the audio representation in order to correct those errors. 
    
    In this section, we describe our proposed methods to integrate the modality confidence information into the deliberation NLU component described in Section \ref{sec:deliberation}. We assume we have the modality confidence score ($\texttt{score}$) ranges between 0 to 1 ($0 \le \texttt{score} \le 1$). A score closer to 1 indicates that the text modality input is highly reliable, while a score closer to 0 indicates that the audio modality input is highly reliable. We will describe how to obtain and encode this modality confidence score in Section \ref{sec:encoder}. 
    
    We explore three different methods to integrate $\texttt{score}$ into E2E NLU component in this study. The first method is multiplication. Given the modality confidence score, we multiply it to text embedding directly, and ($1 - \texttt{score}$) to audio embeddings and then forward to the \texttt{Fusion} module.
    The second method is appending the score to text and audio embeddings then forward to the \texttt{Fusion} module as follows:
    \begin{align}
        e^{\text{MCAT}}_{\text{txt}} &= [e_{\text{txt}}, \texttt{score}]\\
        e^{\text{MCAT}}_{\text{aud}} &= [e_{\text{aud}}, \texttt{score}]\\
        e^{\text{MCAT}}_{\text{fused}} &= \texttt{MHA}_\texttt{fusion}( e^{\text{MCAT}}_{\text{txt}}, e^{\text{MCAT}}_{\text{aud}}, e^{\text{MCAT}}_{\text{aud}})
    \end{align}
    
    Additionally, we can use the score as an additional feature for computing $P_\text{copy}$ in the \texttt{Decoder} module as follows: 
    \begin{align}
        P^{\text{MCAT}}_\text{copy} &= \sigma(\texttt{Linear}([d_v, c_v, \texttt{score}]))
    \end{align}
    In Section \ref{sec:res_incorporate}, we will present the results on the different score integration methods.

    \subsection{Modality Confidence Score Encoder}
    \label{sec:encoder}
    
    \begin{figure}[t]
    \begin{minipage}[b]{1.0\linewidth}
      \centering
      \centerline{\includegraphics[width=6cm]{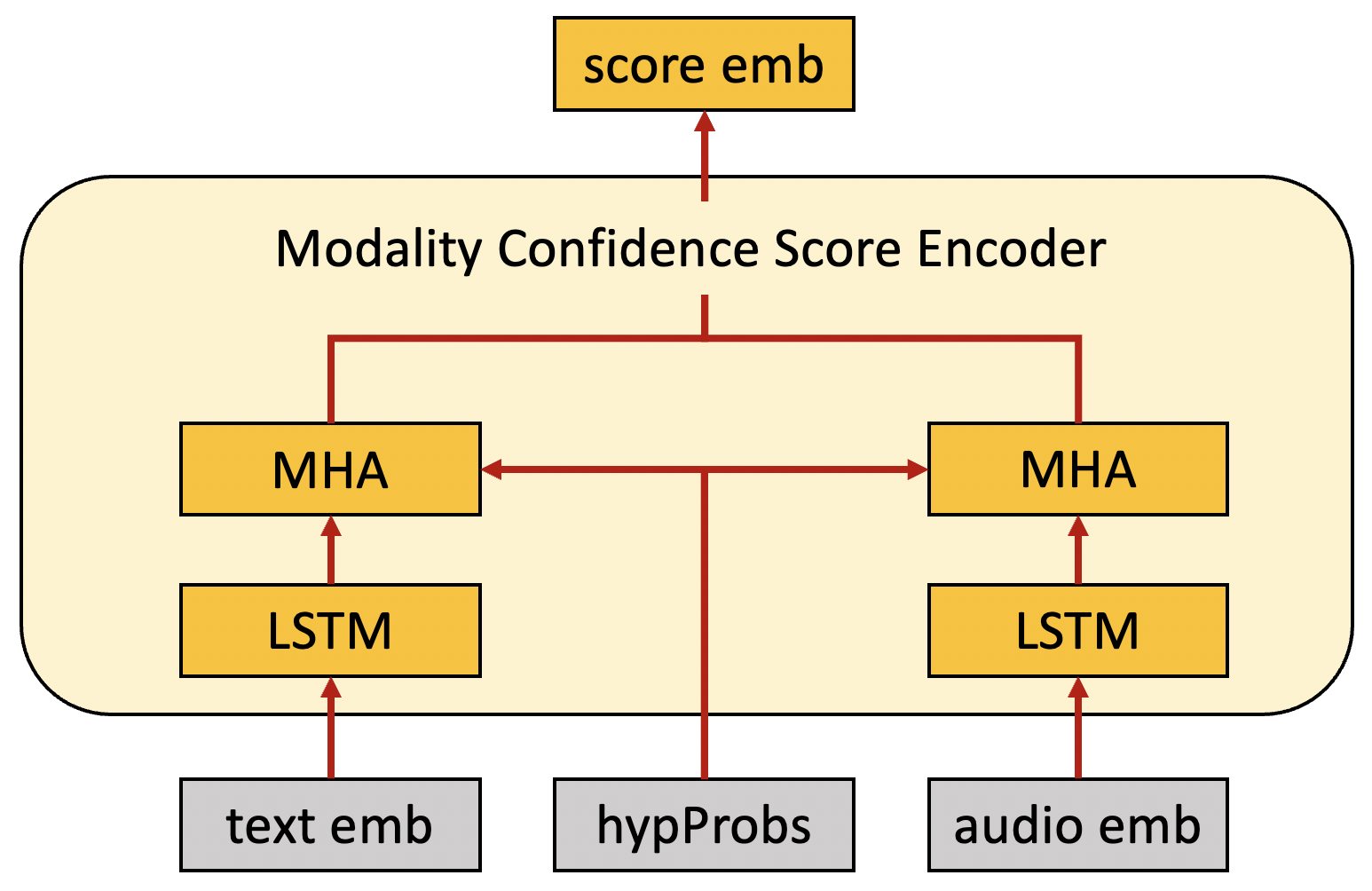}}
    \end{minipage}
    \caption{The overall architecture of our Modality Confidence Score Encoder.}
    \label{fig:score_encoder}
    \end{figure}

    To obtain the modality confidence score (\texttt{score}), we build a Score Encoder (\texttt{ScoreEncoder}) to generate a single score embedding by utilizing all the available information from the frozen RNNT model. This includes the log probability of the ASR hypothesis (\texttt{hypProbs}), text embeddings, and audio embeddings.
    \begin{align}
        \texttt{score} &=  \texttt{ScoreEncoder}(\texttt{hypProbs}, e_\text{txt}, e_\text{aud}) 
    \end{align}
    Since these inputs have different lengths (1, 1:$T$, 1:$U$, respectively), we use a simple \texttt{LSTM} and \texttt{MHA} to map those features into a single score embedding. The \texttt{LSTM} for each modality first takes audio or text embeddings to model sequential information and the \texttt{LSTM} output of each modality is combined with \texttt{hypProbs} using a Multi-Head Attention (\texttt{MHA}). Finally, the \texttt{MHA} output of each modality are concatenated to generate a single score embedding. 
    
    \begin{align*}
        s_\text{txt} &= \texttt{MHA}(\texttt{hypProbs}, \texttt{LSTM}_\text{txt} ( e_\text{txt} ), \texttt{LSTM}_\text{txt} ( e_\text{txt} )) \\
        s_\text{aud} &= \texttt{MHA}(\texttt{hypProbs}, \texttt{LSTM}_\text{aud} ( e_\text{aud} ), \texttt{LSTM}_\text{aud} ( e_\text{aud} )) \\
        \texttt{score} &=  [s_\text{aud}, s_\text{txt}]
    \end{align*}

    Our \texttt{ScoreEncoder} is trained using error information from hypotheses generated by our frozen RNNT model on the STOP training dataset. We use a simple binary target scheme, where 1 represents a correct ASR hypothesis and 0 represents an ASR hypothesis with an error. We first train the \texttt{ScoreEncoder}, followed by joint training of the entire NLU component with the frozen \texttt{ScoreEncoder} and RNNT. We will present the results of different training methods in Section \ref{sec:res_score_encoder}. The overall \texttt{ScoreEncoder} architecture is illustrated in Figure \ref{fig:score_encoder}. We limit the number of parameters in the \texttt{ScoreEncoder} to 0.3M parameters to ensure it is suitable for on-device streaming scenarios.

\vspace{0.3cm}
\section{Experiments}
    \subsection{Data}
    \subsubsection{STOP dataset}
    We used the largest public SLU dataset, STOP (Spoken Task Oriented Semantic Parsing) \cite{stop2022} to evaluate our proposed approach. The STOP dataset is based on Task-Oriented Semantic Parsing (TOPv2) \cite{chen-2020-topv2}, a well-known NLU benchmark, that covers 8 different domains including alarm, messaging, music, navigation, timer, weather, reminder, and event. The spoken data was collected by Amazon Mechanical Turk (MTurk). The dataset split into three subsets: 120k training data, 33k validation data, and 76k evaluation data.   
    
    \subsection{Models}
    \subsubsection{E2E SLU: Frozen ASR Component}
    In this study, which targets on-device streaming use cases, we used three different ASR models with relatively smaller sizes: 10M, 15M, and 25M parameters. All of these models were variants of RNNT, a widely-used architecture in streaming use cases. These models have a 1-layer LSTM predictor, a conformer encoder \cite{gulati2020conformer, Shi2022conformer} with varying numbers of layers (3L, 6L, 13L for 10M, 15M, 25M models, respectively), and a 1-layer of Joiner. We used a 4-stride, 40ms lookahead, and 120ms segment size audio input features, and 4k of sentence piece targets \cite{kudo-richardson-2018-sentencepiece}.
    Using the Alignment Restricted RNN-T loss \cite{Mahadeokar2021AR-RNNT} and SpecAugment techniques \cite{park2019specaugment}, the model was trained on 145k hours of in-house speech data with the same recipe in \cite{le21_interspeech}. 
    The Word Error Rates (WER) for each split of the STOP datasets, using ASRs of three different sizes (10M, 15M, and 25M), are presented in Table \ref{tab:asr}.  
    
    \begin{table}[h]
        \caption{The WER results for each split of the STOP datasets, using the frozen RNNT ASR of three different sizes (10M, 15M, and 25M).}
        \centering
        \begin{tabular}{rrrr}
        \toprule
                  & \multicolumn{3}{c}{\textbf{WER}} \\
        \textbf{Frozen ASR (RNNT)} & \textbf{train}  & \textbf{valid}  & \textbf{test}  \\
        \midrule
        10M               & 7.68    & 6.99  & 6.54  \\
        15M               & 5.36    & 4.88  & 4.59  \\
        25M               & 4.19    & 3.83  & 3.54  \\
        \bottomrule
        \end{tabular}
    \label{tab:asr}
    \end{table}
    
    Note that the ASRs were kept frozen and were not fine-tuned with the NLU component for focusing on improving NLU component while maintaining the table transcription performance of the ASRs.  
    
    \subsubsection{E2E SLU: NLU Component}
    For NLU component for on-device streaming use cases, we used 5M parameters of deliberation-based NLU architecture as described in Section \ref{sec:deliberation}. From the frozen ASR, 256 dimensional audio and text embeddings were passed into the \texttt{Fusion} module. The \texttt{Fusion} module is Multi-Head Attention (\texttt{MHA}) with 8 attention heads and the fused features are then passed into the \texttt{Pooling} module consists of 2 transformer \cite{vaswani2017attention} encoder layers with 8 attention heads. The output feature of \texttt{Pooling} is 224 dimensional feature and then finally passed into the \texttt{Decoder} module consistent of a single transformer decoder layer with 2 attention heads with a pointer-generator network with 1 attention head \cite{decoupled}. We used 586 ontology tokens including semantic parse in addition to 4k sentence pieces that used in ASR. 
    The baseline NLU component was trained on STOP dataset with using \texttt{union} strategy, a combination of reference text and hypothesis from ASR with the same recipe in \cite{le21_interspeech}. 
    
    For our \texttt{ScoreEncoder}, we used a single layer of LSTM with 128 cells for each modality. Initially, we trained the model and subsequently kept it frozen without fine-tuning it with the NLU component.

    Our E2E SLU models were evaluated using Exact Match (EM)~\cite{chen-2020-topv2}, which measures the accuracy of the model's hypothesis by comparing it to the reference annotation using a string match, while ignoring punctuation and casing. Both the parse structure and slot content transcription need to be matched to be considered correct.

\vspace{0.3cm}
\section{Results and Analysis}
\label{sec:res}

    \subsection{How to Incorporate the Modality Confidence Information?}
    \label{sec:res_incorporate}
    We first investigated the effective way to integrate the confidence information into the deliberation-based NLU component. We compared three different methods to integrate the confidence information: (1) multiplication in the fusion module, (2) appending in the fusion module, and (3) appending in both fusion and decoder module (described in Section \ref{sec:integration}). In this experiment, we used oracle modality confidence score that we defined based on WER as follows:
    \begin{align}
        \texttt{score}^{oracle} &= 1.0 - min(1.0, \texttt{WER}) 
    \end{align}
    Note that $\texttt{score}^{oracle}$ ranges between 0 to 1, and 1 represents a correct ASR hypothesis and 0 represents a hypothesis with an ASR error.
    As shown in Table \ref{tab:integration}, all three methods show higher EMs for both the w/ ASR error and w/o ASR error cases compared to the baseline E2E SLU. We also observed similar gains in EM when using either appending or multiplication method in fusion module. However, we found a significant increase in EM by appending method in decoder module, particularly in ASR error cases. 
    
    \begin{table}[h]
        \caption{The EM results for the baseline E2E SLU and three different integration methods of the modality confidence information: (1) MCAT with multiplication method in the fusion module (MUL FUSION), (2) MCAT with appending method in the fusion module (APPEND FUSION), and (3) MCAT with appending method in both fusion (APPEND FUSION) and decoding modules (APPEND DEC). }
        \centering
        \begin{tabular}{lrr}
        \toprule
        \textbf{Integration Methods}  & \multicolumn{2}{c} {\textbf{ASR Error}} \\
                              &    Yes (61k utters)               &  No (14k utters) \\
        \midrule
        Our Baseline             & 84.4 & 32.3   \\
        \quad + MUL FUSION          & 84.7 & 35.5  \\
        \quad + APPEND FUSION       & \textbf{84.9} & 35.4  \\
        \quad \quad + APPEND DEC & 83.7 & \textbf{41.8}  \\ 
        \bottomrule
        \end{tabular}
    \label{tab:integration}
    \end{table}

    \subsection{Building a Modality Confidence Score Encoder}
    \label{sec:res_score_encoder}
    We next investigated what is the best way to encode the modality confidence information. 
    We experiment different strategies to build Modality Confidence Score Encoder. We first observed that using three input resources performed the best (1) text embedding from ASR RNNT predictor, (2) audio embedding from ASR RNNT encoder, and (3) ASR hypothesis probability (\texttt{HypProb}). 
    For the objective function to train the Modality Confidence Score Encoder, we tried several options such as classification with weighted class, regression, and focal loss \cite{lin2017focal}. We found that the binary weighted classification performed the best results. In our experiments, we assigned a class weight of 0.3 to the ``1'' label, which represents a correct ASR hypothesis, and a class weight of 0.7 to the ``0'' label. 
    
    One of our main challenges in building the Modality Confidence Score Encoder was dealing with unbalanced training data, as our ASR system achieved the high accuracy on the STOP dataset (as shown in Table \ref{tab:asr}). For example, the target of training data was heavily skewed, the majority of examples having a score of 1. To resolve this problem, we addressed the class imbalance in the training dataset by augmenting the examples with ASR errors (``0'' label). We added noise to the audio of the original STOP training data by using Noise Injection technique \cite{reddy2021interspeech} and generated examples with ASR errors intentionally. 

    \subsection{How good should Modality Confidence Score Encoder perform to improve EM?} 
\begin{figure}[h]
    \begin{minipage}[b]{1.0\linewidth}
      \centering
      \centerline{\includegraphics[width=7.5cm]{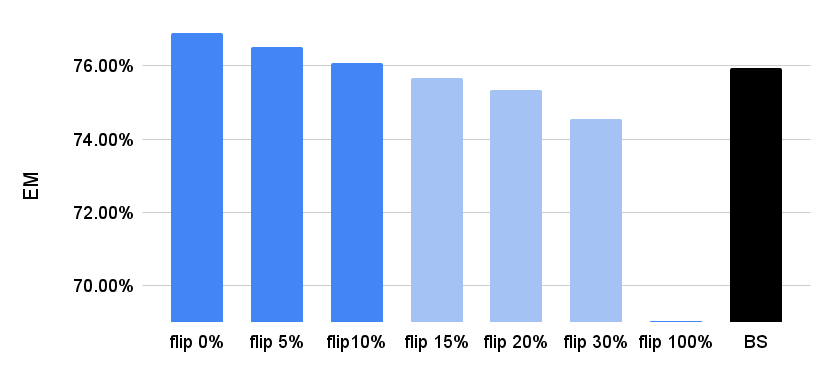}}
    \end{minipage}
    \caption{The EM results of the baseline (BS) and our proposed approach MCAT with different flipping ratios (ranging from 0\% to 100\%) in corrupted scores. "flip 0\%" indicates no flipping, while "flip 100\%" indicates 100\% flipping.  }
    \label{fig:flip}
    \end{figure}

    We analyzed the minimum performance requirement for the Score Encoder to improve EM. During training the NLU module, we used the binary oracle score and then intentionally introduced errors during decoding by randomly flipping the oracle score (1 $\Longleftrightarrow$ 0) in order to simulate estimation from the imperfect Score Encoder. Figure \ref{fig:flip} shows he EM results of the baseline and our proposed approach MCAT with different flipping ratios (ranging from 0 to 100\%) in corrupted scores. The results showed that the Score Encoder needs to achieve at least 87\% accuracy to improve EM compared to the baseline. Interestingly, we observed that when the score was flipped 100\% of the time, the model performed significantly worse than the baseline. These result indicates that our MCAT approach functions as intended and its effectiveness is not due to randomness.
    
    \subsection{Semantic Parsing Results (EM)}
    Table \ref{tab:em_model_size} shows the EM of (1) the baseline E2E SLU, (2) E2E SLU with Oracle Confidence Score, and (3) our proposed approach, Modality Confidence Aware Training (MCAT) on STOP dataset with varying sizes of the frozen ASR RNNT models (10M, 15M, and 25M parameters). We observed an absolute EM improvement of 0.41, 0.35, and 0.11 with our proposed method when using ASR models of size 10M, 15M, and 25M parameters, respectively. These results suggest that our MCAT incorporating modality confidence information may be particularly beneficial when the ASR model is less accurate, such as on-device streaming use cases. On the other hand, as the ASR performance improves, the potential benefits of using our MCAT may be diminished.      

    \begin{table}[h]
        \caption{The EM results for three different models: (1) Baseline E2E SLU, (2) Baseline with Oracle Confidence Info, and (3) our proposed model, MCAT, on the STOP dataset with varying sizes of ASR RNNT models (10M, 15M, and 25M parameters), along with the same size of the NLU component (5M parameters).}
        \centering
        \begin{tabular}{rrrr}
        \toprule
         \textbf{NLU}                         & \multicolumn{3}{r} {\textbf{Frozen ASR (RNNT)}}   \\
         5M                      & 10M       & 15M       & 25M       \\
        \midrule
        Our baseline E2E SLU & 68.37     & 71.97     & 74.05     \\
        w/ oracle confidence Info & 69.66     & 72.95     & 74.90     \\
        w/ our MCAT               & 68.78     & 72.32     & 74.16     \\
        \bottomrule
        \end{tabular}
        \label{tab:em_model_size}
    \end{table}

\vspace{0.3cm}
\section{Conclusions}
\label{sec:conclusions}
    We have introduced our novel approach for building robust end-to-end (E2E) spoken language understanding (SLU) models leverages modality confidence information to intelligently fuse audio and text input representations in the NLU component. The model is designed to prioritize the audio representation when the quality of the text representation is poor due to ASR hypothesis errors. In our experiments on the public STOP dataset with an on-device streaming scenario, our approach MCAT outperformed strong E2E models. Going forward, we plan to extend our method by incorporating token-based modality confidence information and exploring its effectiveness on other datasets with different modalities and scenarios.

\bibliographystyle{IEEEtran}
\bibliography{refs}

\end{document}